\pdfoutput=1

\documentclass[11pt]{article}

\usepackage{EMNLP2023}

\usepackage{times}
\usepackage{latexsym}
\usepackage{placeins}

\usepackage[T1]{fontenc}

\usepackage[utf8]{inputenc}

\usepackage{microtype}

\usepackage{inconsolata}

\usepackage{amsmath}
\usepackage{amsfonts}
\usepackage{hhline}
\usepackage{multirow}

\usepackage[T1]{fontenc}    
\usepackage{url}            
\usepackage{booktabs, multirow, tabularx}       
\usepackage{amsfonts}       
\usepackage{nicefrac}       
\usepackage{microtype}      
\usepackage{graphicx}
\usepackage{natbib}
\usepackage{doi}
\usepackage{color}
\usepackage{amsmath}
\usepackage{mathtools, cuted}
\usepackage{etoolbox}
\usepackage{changepage}
\usepackage{pdflscape}
\usepackage{multicol}
\usepackage{balance}
\usepackage{wrapfig}
\usepackage{array}
\usepackage{soul}
\usepackage{bm}
\usepackage{enumitem}
\usepackage[symbol]{footmisc}

\AfterEndEnvironment{strip}{\leavevmode}

\newcommand{\nce}{\mathrm{NCE}}

\newcommand{\JEmbeddingVTwo}{\texttt{jina-embeddings-v2}}
\newcommand{\JEmbeddingVThree}{\href{https://huggingface.co/jinaai/jina-embeddings-v3}{\texttt{jina-embeddings-v3}}}

\newcommand{\rom}[1]{\uppercase\expandafter{\romannumeral #1\relax}}

\def\@fnsymbol#1{\ensuremath{\ifcase#1\or *\or \dagger\or \ddagger\or
   \mathsection\or \mathparagraph\or \|\or **\or \dagger\dagger
   \or \ddagger\ddagger \else\@ctrerr\fi}}
\newcommand{\ssymbol}[1]{^{\@fnsymbol{#1}}}

\definecolor{darkgreen}{rgb}{0.0, 0.5, 0.1}

\title{\JEmbeddingVThree: Multilingual Embeddings With Task LoRA}

\author{Saba Sturua$^*$, Isabelle Mohr$^*$, Mohammad Kalim Akram$^*$ \\ \textbf{Michael G\"unther}$^*$, \textbf{Bo Wang}$^*$, \textbf{Markus Krimmel}, \textbf{Feng Wang} \\ \textbf{Georgios Mastrapas}, \textbf{Andreas Koukounas}, \textbf{Nan Wang} \and \textbf{Han Xiao} \\
Jina AI GmbH, Prinzessinnenstraße 19--20, 10969 Berlin, Germany \\
\texttt{research@jina.ai}
}

\hypersetup{
pdftitle={Jina Embeddings V3: Multilingual Embeddings With Task LoRA},
pdfauthor={Saba Sturua, Isabelle Mohr, Mohammad Kalim Akram, Michael G\"unther, Bo Wang, Markus Krimmel, Feng Wang, Georgios Mastrapas, Andreas Koukounas, Nan Wang, Han Xiao},
pdfkeywords={Embeddings, Multilingual Models, Token length, Semantic Textual Similarity, Information Retrieval, Text Retrieval},
}

\begin{document}
\maketitle
\def\thefootnote{*}\footnotetext{Equal contribution.}\def\thefootnote{\arabic{footnote}}

\begin{abstract}
We introduce \JEmbeddingVThree{}, a novel text embedding model with 570 million parameters, achieves state-of-the-art performance on multilingual data and long-context retrieval tasks, supporting context lengths of up to 8192 tokens. The model includes a set of task-specific Low-Rank Adaptation (LoRA) adapters to generate high-quality embeddings for query-document retrieval, clustering, classification, and text matching. Evaluation on the MTEB benchmark shows that \JEmbeddingVThree{} outperforms the latest proprietary embeddings from OpenAI and Cohere on English tasks, while achieving superior performance compared to \texttt{multilingual-e5-large-instruct} across all multilingual tasks. With a default output dimension of 1024, users can flexibly reduce the embedding dimensions to as low as 32 without compromising performance, enabled by Matryoshka Representation Learning.
\end{abstract}

\section{Introduction}
\label{sec:introduction}

Text embedding models represent documents as high-dimensional vectors, converting semantic relationships between documents into spatial relationships between vectors. These models are fundamental to neural information retrieval and have been widely adopted across various domains of NLP and IR research and applications. Text embeddings are utilized in diverse downstream tasks such as classification, retrieval, and clustering. Notably, they have gained significant traction in building Retrieval-Augmented Generation (RAG) systems, where they serve as the primary technique in the retrieval step.

A major limitation of traditional embedding models is that, despite being named as general-purpose, they often require fine-tuning for specific tasks~\cite{jiao2020tinybert} and frequently struggle with common failure cases~\cite{gao2021simcse}. To address this, recent research has increasingly focused on leveraging large language models (LLMs) as the backbone for general-purpose embedding generation, capitalizing on their ability to efficiently handle multiple languages and tasks~\cite{jiang2024scaling}. However, with model sizes typically reaching 7 billion parameters, deploying these models in real-world applications poses significant challenges. Furthermore, the marginal improvements in evaluation metrics offered by LLM-based embeddings, compared to encoder-only embedding models, render them a less practical choice for many use cases.

This paper introduces \JEmbeddingVThree{}, a novel text embedding model with 570 million parameters, optimized for multilingual data, long-context retrieval, and high performance across multiple tasks. Evaluation on the MTEB benchmark demonstrates that \JEmbeddingVThree{} not only significantly improves upon its predecessor, \JEmbeddingVTwo{}~\cite{gunther2023jina2} and its bilingual variants~\cite{mohr2024multi}, but also outperforms the latest proprietary embeddings from OpenAI and Cohere on English tasks, while surpassing \texttt{multilingual-e5-large-instruct} across all multilingual tasks. Additionally, compared to LLM-based embeddings such as \texttt{e5-mistral-7b-instruct}, which has a parameter size of 7.1 billion (12x larger) and an output dimension of 4096 (4x larger) but offers only a 1\% improvement on MTEB English tasks, \JEmbeddingVThree{} is a far more cost-efficient solution, making it more suitable for production and on-edge computing. The key contributions of this paper are:

\begin{itemize}
    \item \textbf{Task-specific optimization with LoRA:} We demonstrate that LoRA adapters~\cite{hulora} effectively generate task-specific embeddings, outperforming prior instruction-based approaches.

    \item \textbf{Patching retrieval failures with synthetic data:} A qualitative analysis identified four common types of retrieval failures. We mitigated these issues by incorporating synthetic training data, thereby improving model robustness on edge cases.

    \item \textbf{Integration of latest techniques:} Our model incorporates several key advancements, including Matryoshka Representation Learning~\cite{kusupati2022matryoshka}, instruction tuning~\cite{wei2022finetunedlanguagemodelszeroshot,su2023one}, and long-context retrieval~\cite{gunther2023jina2}.
\end{itemize}

Section~\ref{sec:related_work} provides an overview of prior research relevant to the objectives of this paper.
Section~\ref{sec:model-architecture} presents the architecture of \JEmbeddingVThree{} in detail.
The training procedure is described in Section~\ref{sec:backbone_pretraining}.
In Section~\ref{sec:evaluation}, we conduct a thorough multilingual evaluation, including ablation studies that offer insights into the impact of our architectural and training decisions.

\section{Related Work}
\label{sec:related_work}
\subsection{General Text Embeddings}

In recent years, significant progress has been made in the field of text embeddings, largely driven by the emergence of transformer-based pre-trained language models that capture the underlying semantics of language effectively~\cite{devlin2019bert}. However, these models are predominantly trained with a masked language modeling (MLM) objective, which is not optimal for generating high-quality text embeddings. To overcome this limitation, recent approaches have focused on fine-tuning and extending these models specifically for embedding tasks~\cite{reimers2019sentence}.

A key advancement in this area is the development of multi-stage and multi-task fine-tuning strategies that incorporate weakly-supervised contrastive training~\cite{wang2022text,gunther2023jina2,mohr2024multi}. These methods improve the versatility of embeddings, enabling models to perform well across a diverse range of applications and tasks, as opposed to models trained solely on semantic textual similarity datasets.

Furthermore, techniques such as AliBi~\cite{press2022alibi} and RoPE~\cite{su2024roformer} have enabled models like \JEmbeddingVTwo{}~\cite{gunther2023jina2} to handle longer sequences, up to 8192 tokens, by replacing absolute positional encoding with relative encoding methods. To make embeddings more compact, Matryoshka Representational Learning (MRL)~\cite{kusupati2022matryoshka} enables the truncation of embeddings without compromising performance on downstream tasks by modifying the loss function used during training.

\subsection{Multilingual Embedding Models}
One of the earliest multilingual transformer models is Multilingual BERT (mBERT)~\cite{devlin2019bert}, trained on 104 languages. This was followed by XLM~\cite{conneau2019xlm} and XLM-RoBERTa (XLM-R)~\cite{conneau2020xlmr}, which utilized parallel data during training. \citet{wang2024multilingual} extends this work by fine-tuning XLM-R on high-quality multilingual labeled datasets and applying knowledge distillation from a cross-encoder to further improve embedding quality. Similarly, \citet{chen2024bge} introduced BGE M3, another XLM-R-based model that supports longer sequences. The authors extended XLM-R’s maximum sequence length to 8192 tokens, continued pre-training with the RetroMAE method~\cite{xiao2022retromae}, and fine-tuned it contrastively using a novel multi-CLS pooling strategy. mGTE~\cite{zhang2024mgte} also builds on XLM-R, incorporating RoPE positional embeddings~\cite{su2024roformer}.

Another approach leverages LLMs for multilingual embeddings~\cite{zhang2023language,wang2023improving}, benefiting from their extensive language support and diverse training data. However, LLMs are computationally inefficient due to their larger size, making them less practical for many applications. To address this, \citet{lee2024gecko} generate and relabel training data to distill knowledge from LLMs into a compact encoder model, avoiding the need for direct fine-tuning of the larger LLMs.

\subsection{Task-Specific Embedding Models}
\label{sec:rel:task-specific}
Previous research has highlighted limitations in training models to produce generic embedding vectors that perform well across various use cases and domains. For example, \citet{wang2022text} observed that in asymmetric retrieval tasks, such as question answering and typical information retrieval, models perform better by appending distinct prefixes to queries and documents before encoding. While the E5 models from this work employ a single prefix for all queries and another for all documents, \citet{su2023one} introduced more complex instructions to encode additional information about relevance in retrieval tasks and the domain of the data.

\citet{hulora} propose a technique that uses lightweight LoRA layers to fine-tune LLMs. By freezing the original model weights, this approach significantly improves training efficiency. More importantly, deploying multiple fine-tuned instances becomes feasible, as the LoRA typically require less than 1\% of the memory needed for the original model weights. However, to the best of our knowledge, this technique has not yet been explored as an alternative to instruction-based methods in embedding training.

\section{Model Architecture}
\label{sec:model-architecture}
\begin{figure*}[htbp]
\centering
\includegraphics[width=\linewidth]{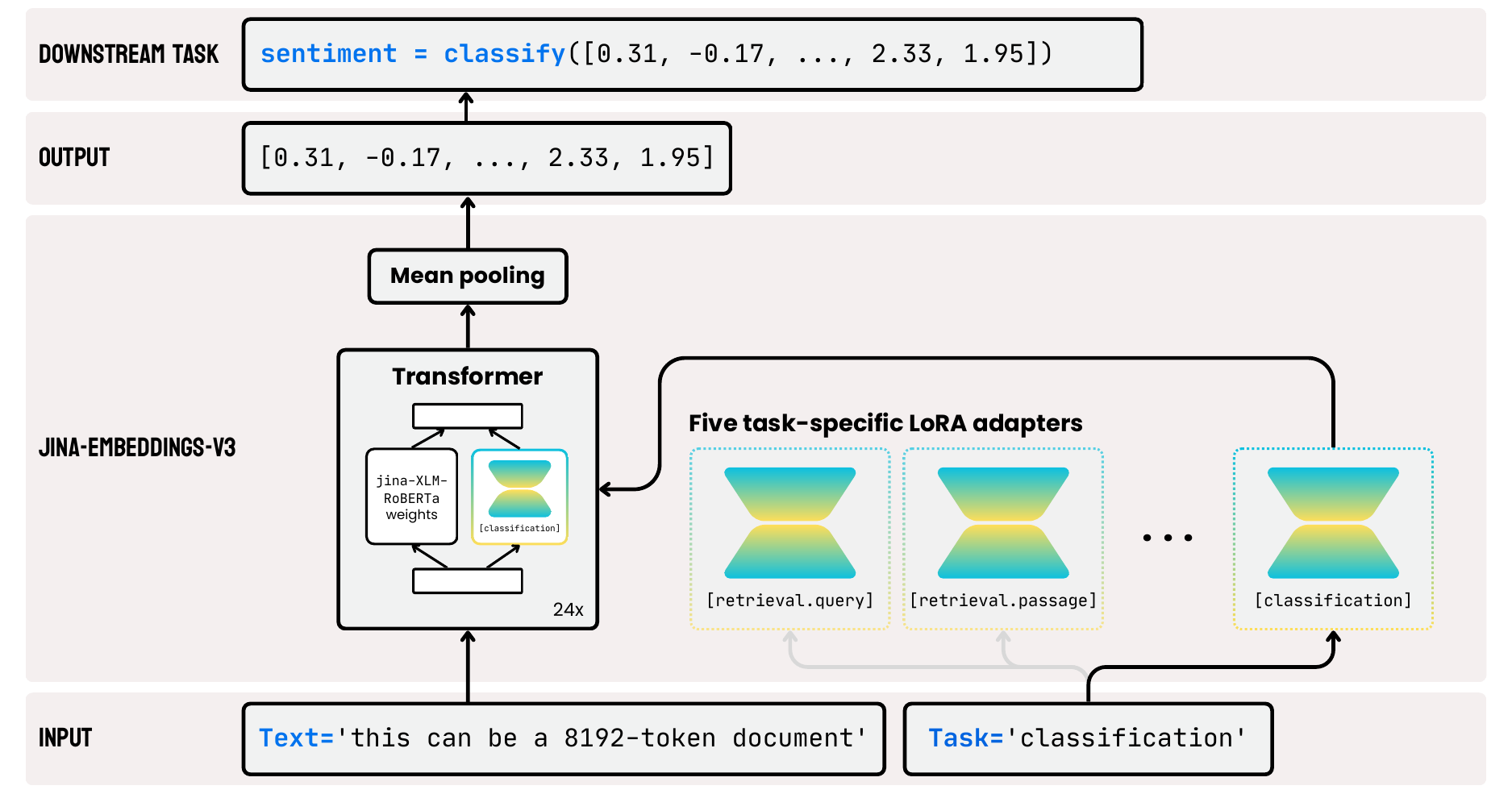}
\caption{The architecture of \JEmbeddingVThree{} is based on the XLM-RoBERTa model, with several key modifications. FlashAttention 2 is integrated for enhanced computational efficiency, while RoPE extends support for sequences up to 8192 tokens. Task-specific LoRA adapters are introduced to optimize embeddings for various tasks. The model’s input consists of two parts: the text, which is the long document to be embedded, and the task type. \JEmbeddingVThree{} supports four tasks and implements five adapters to choose from: \texttt{retrieval.query} and \texttt{retrieval.passage} for query and passage embeddings in asymmetric retrieval tasks, \texttt{separation} for clustering and reranking tasks, \texttt{classification} for classification tasks, and \texttt{text-matching} for tasks involving semantic similarity, such as STS or symmetric retrieval.}
\label{img:model-architecture}
\end{figure*}
The architecture of \JEmbeddingVThree{} is depicted in Figure~\ref{img:model-architecture}. To implement the backbone architecture, we adapt the XLM-RoBERTa model with modifications that (1) enable effective encoding of long text sequences, (2) allow task-specific encoding of embeddings, and (3) increase model efficiency. \JEmbeddingVThree{} retains the original XLM-RoBERTa tokenizer.

As outlined in Table~\ref{tab:parameter-count}, \JEmbeddingVThree{} is larger than \JEmbeddingVTwo{}, but significantly smaller than embedding models fine-tuned from LLMs~\cite{lee2024nv,wei2022finetunedlanguagemodelszeroshot}. Importantly, the LoRA adapters account for less than 3\% of the total parameters, adding minimal overhead. To further enhance performance and reduce memory consumption, we utilize FlashAttention 2~\cite{daoflashattention}, support activation checkpointing, and employ the DeepSpeed framework~\cite{rasley2020deepspeed} for efficient distributed training.

\begin{table*}[htb]
\centering
\small{
\begin{tabular}{cccccccc}
\toprule
 \multicolumn{2}{c}{Parameters} & Max input tokens & Max output dim. & Layers & Vocabulary & Attention Impl. & Pooling\\
\cmidrule(lr){1-2}
 Base & w/ adapters &&  && & &\\
\midrule
559M & 572M & 8192 & 1024$^*$ & 24 &  250K & FlashAttention2 & Mean\\
\bottomrule
\end{tabular}
}
\caption{Model specification of \JEmbeddingVThree{}. The maximum output dimension is 1024, but with Matryoshka Representation Learning (MRL), users can choose any dimension lower than 1024, such as 16 or 32, allowing for a trade-off between space efficiency and performance. This trade-off is studied in Table~\ref{tab:MRL_averages}.}
\label{tab:parameter-count}
\end{table*}

To handle long text sequences, we replace absolute positional embeddings with Rotary Position Embeddings (RoPE)~\cite{su2024roformer}, which use a rotation matrix to encode absolute positions while embedding relative positional dependencies directly within the self-attention mechanism. We also experimented with extending positional encodings~\footnote{\url{https://spaces.ac.cn/archives/7947} (Accessed 08-27-2024)} as done in the BGE M3 model~\cite{bge-m3}, but observed poor performance on tasks involving long texts. This could be attributed to differences in training data and pooling strategies, as we trained primarily on short texts and used mean pooling instead of multi-CLS pooling.

\citet{xiong2024effective} demonstrated that increasing the base frequency parameter of rotary positional embeddings enhances performance on long-text tasks, while \citet{zhang2024mgte} adjusted the rotary base frequency during training on short sequences to generalize better on longer sequences. We found that setting a rotary base frequency of 10,000 during training and adjusting it to 20,000 during inference improves performance on long-text tasks without degrading performance on short-text tasks.

Embeddings can be used for various downstream tasks, including clustering, retrieval, and classification, each requiring different interpretations of the representation space resulting in different similarity metrics. Asymmetric retrieval tasks, for instance, benefit from encoding queries and documents differently. \citet{wang2022text} suggest that using distinct instructions for queries and documents improves the effectiveness of embedding models in such tasks.

However, writing effective task-specific instructions is non-trivial. As an alternative, we employ task-specific LoRA adapters. The embedding and linear layers within the multi-head attention mechanism are equipped with low-rank decomposition matrices of rank 4. These task-specific LoRA adapters are loaded alongside the model weights and are dynamically selected based on the input task type. Each text input in the batch is associated with a task descriptor, represented as an integer corresponding to the LoRA adapter's ID.

\section{Training Method}
\label{sec:backbone_pretraining}
%
We initialize the model using the weights of the original XLM-RoBERTa model. However, the model's original MLM objective is not fully aligned with our training objectives due to the changes in positional embedding methods. Despite this, we observe that initializing with pre-trained weights leads to faster convergence during pre-training compared to random initialization.

Our training paradigm consists of three stages, as is common for training text embedding models:

\begin{enumerate}[label=\Roman*]
    \item \textbf{Pre-Training:}
    We perform standard MLM training using large multilingual text corpora. The model is initialized with XLM-RoBERTa weights to expedite pre-training and avoid training from scratch.\label{stage:one}

    \item \textbf{Fine-Tuning for Embedding Tasks:} To learn how to encode a text passage into a single vector representation, we follow the approach outlined in~\cite{gunther2023jina2}. This method incorporates a pooling layer into the transformer model to aggregate token representations into a single embedding vector and fine-tunes the model on pairs of semantically related texts.\label{stage:two}

    \item \textbf{Training Task-Specific Adapters:} We train five LoRA adapters for four different tasks using dedicated datasets and task-specific loss functions to optimize performance for each use case.\label{stage:three}
\end{enumerate}
\subsection{Pre-Training}

After initialization, the model is trained using the MLM objective with whole word masking~\cite{devlin2019bert}. At this stage, we only train the transformer model, excluding the LoRA adapters and pooling layer.

To support multilingual tasks, the training data is drawn from the CulturaX corpus~\cite{nguyen2023culturax}, which includes data from 89 languages, with English comprising approximately 20\% of the dataset. During training, each batch contains data for only a single language, but we rotate the language between batches.

For long-context support, we first train for $100,000$ steps on text sequences that are truncated to 512 tokens, followed by an additional $60,000$ steps using on text sequences truncated to 8192 tokens and a reduced batch size. The details are provided in Appendix~\ref{table:hyperparameters}.

To enhance the model's ability to represent long text documents, we train with a low rotary base value, increasing it during inference as described in Section~\ref{tab:parameter-count}. However, we observed that the model's ability to encode long documents still lagged behind models such as \JEmbeddingVTwo{}. To address this, we extended the training on long-text data, which resulted in improved performance on long-text retrieval tasks such as NarrativeQA. See Section~\ref{ssec:longembed} for further details.

\subsection{Fine-Tuning for the Embedding Task}
\label{sec:embedding-training}
After pre-training, we fine-tune the model to encode a text sequence into a single vector representation. Following the Sentence-BERT approach~\cite{reimers2019sentence}, we augment the model with a mean pooling layer to aggregate the semantics from all output token vectors into a single vector representation. The fine-tuning procedure follows \citet{mohr2024multi}, where the model is trained on text pairs using a bi-directional InfoNCE~\citep{DBLP:journals/corr/abs-1807-03748} loss, $\mathcal{L}_{\mathrm{pairs}}$:

\begin{equation}
    \mathcal{L}_{\mathrm{pairs}}(B) := \mathcal{L}_{\nce}(B) + \mathcal{L}_{\nce}(B^\dagger)
\end{equation}
defined on a batch $B = ((p_1, q_1),\ldots,(p_k, q_k))$ of $k$ pairs, and $B^\dagger= ((q_1, p_1),\ldots,(q_k, p_k))$ (obtained from $B$ by swapping the order of pairs). $\mathcal{L}_{\nce}$ denotes the following loss function:
\begin{equation}
    \mathcal{L}_{\nce}(B) := -\sum_{(x_i, y_i) \in B} \ln \frac{e^{s(x_i, y_i) / \tau}}{\sum\limits_{i' = 1}^k e^{s(x_i, y_{i'}) / \tau}}
\end{equation}
The training data consists of over one billion text pairs, drawn from more than 300 distinct sub-datasets, each representing specific domains in various languages. During training, the data loader constructs each batch by sampling a specific sub-dataset, ensuring that only text pairs from that dataset and language are included.

For data preparation, we follow the same methodology as our previous work~\cite{mohr2024multi}, with an additional filtering step. This filter removes pairs where at least 80\% of the words (minimum of four) in the shorter text are sub-strings of the longer text. This filtering step increases the difficulty of the training and encourages the model to focus less on syntactic overlap.

As in the pre-training phase, we begin training on short text pairs, followed by further training on longer texts using a larger sequence length but a reduced batch size. In this phase, we use only a subset of the datasets containing sufficiently long texts.

\subsection{Training Task-Specific Adapters}
\label{sec:supervised_fine_tuning}
Related work on embedding training~\cite{wang2022text, xiao2023c} introduces an additional generic training phase following the pair-wise training phase. This phase incorporates high-quality data from various task types to optimize model performance across a range of downstream use cases. In this stage, recent approaches use task-specific instructions to help the model distinguish between different tasks and domains, as discussed in Section~\ref{sec:rel:task-specific}.

However, this approach complicates the usage, as users must learn task-specific instructions (i.e., prompts) that align with the model's behavior or "vibe." While this offers flexibility, it also makes the model's behavior harder to predict. In contrast, we train five distinct LoRA adapters for four well-defined task types as defined in Table~\ref{tab:task-value}. These tasks are trained independently, with the base model's weights kept frozen. For query-document retrieval, two adapters are trained \emph{jointly}: one for queries and one for passages. During inference, users can select the appropriate adapter based on their downstream task and input role, ensuring optimal embeddings for their specific use case.

\begin{table}[h!]
\centering
\small{
\begin{tabular}{lp{4cm}}
\toprule
\texttt{task} value      & Task Description \\
\midrule
\texttt{retrieval.passage}        & Embedding \textbf{documents} in a query-document retrieval task \\
\texttt{retrieval.query}          & Embedding \textbf{queries} in a query-document retrieval task \\
\texttt{separation}               & Clustering documents, visualizing a corpus \\
\texttt{classification}           & Text classification \\
\texttt{text-matching}            &  Semantic text similarity, general symmetric retrieval, recommendation, finding similar items, deduplication \\
\bottomrule
\end{tabular}
}
\caption{Supported tasks of \JEmbeddingVThree{}, each corresponding to a LoRA adapter and trained independently, except for \texttt{retrieval.passage} and \texttt{retrieval.query}, which are trained jointly.}
\label{tab:task-value}
\end{table}

\subsubsection{Classification Adapter}
The classification adapter generates embeddings that are effective for training downstream classification models, particularly logistic regression classifiers. To train the adapter, we employ the classification training method proposed for the Gecko embedding model~\cite{lee2024gecko}. Specifically, we select datasets covering a range of common classification tasks, including sentiment analysis, intent classification, and article categorization.

From each dataset, we construct tuples consisting of two text values from the same class $(q, p)$ and seven text values from different classes $(n_1, \dots, n_7)$, resulting in a tuple of nine text values $(q, p, n_1, \dots, n_7)$. The model is trained to assign a high cosine similarity to the embeddings of $q$ and $p$, while enforcing low cosine similarity between $q$ and all $n_i$. Each batch is composed of tuples sampled from a single dataset.

We employ an extended version of the InfoNCE loss $\mathcal{L}_{\text{triplet}}$ described in our previous work~\cite{gunther2023jina2} to take into account these additional negative samples.
\begin{flalign}
&\mathcal{L}_{\text{triplet}}(B) := \nonumber\\
&\;\;\;\;\;\mathbb{E}_{r\sim B}\Bigg[-\ln \frac{e^{s(q, p) / \tau}}{\sum\limits_{i = 1}^k \Big[ e^{s(q, p_i) / \tau}+ \sum\limits_{j = 1}^{m} e^{s(q, n_{j,i}) / \tau}\Big]}\Bigg]\nonumber \\
&\, + \mathbb{E}_{r\sim B}\Bigg[-\ln \frac{e^{s(p, q) / \tau}}{\sum\limits_{i = 1}^k e^{s(p, q_i) / \tau}}\Bigg]\nonumber \\
&\text{with}\; r = (q,p, n_1, \ldots, n_{m}).\label{eq:loss-multi-negatives}
\end{flalign}
When using this loss function, texts from the same class as text $q_i$ that occur in the same batch as negatives for a different text $q_j$ $(i \neq j)$ are also treated as negatives. This introduces the problem of false negatives.

To address this, \citet{lee2024gecko} propose appending a unique ID specific to each tuple $r$ to its corresponding text values. This allows the model to easily differentiate between text values from different tuples, enabling it to focus on separating text values within the same tuple in a batch.

\subsubsection{Text Matching Adapter}
\label{sec:text-matching}
This adapter is trained to produce embeddings that quantify the similarity between two text values. It is applicable for tasks such as semantic textual similarity (STS) and retrieval tasks where there is no clear distinction between query and target text values. An example of such a retrieval task is duplicate detection, where text values from a corpus are compared against each other. In these cases, the ``query'' and ``corpus'' texts are treated symmetrically.

To train this adapter, we use the CoSent loss function:
$\mathcal{L}_\mathrm{co}$\footnote{\url{https://github.com/bojone/CoSENT} (Accessed: 08-28-2024)}, as previously employed by \citet{li2024aoe}:

\begin{flalign}
\label{eq:cosentloss}
    &\mathcal{L}_{\mathrm{co}}(B) := \ln \Big[  1 + \sum\limits_{\substack{(q_1,p_1), \\ (q_2,p_2) \in B}} \frac{e^{s(q_2,p_2)} - e^{s(q_1,p_1)}}{\tau} \Big] && \nonumber \\
    &\text{where}\; \zeta(q_1, p_1) > \zeta(q_2, p_2) &&
\end{flalign}

The CoSent loss operates on two pairs of text values, $(q_1, p_1)$ and $(q_2, p_2)$, which are constructed from the batch by selecting combinations of four text values where the ground truth similarity $\zeta$ is provided in the training dataset, and $\zeta(q_1, p_1)$ is greater than $\zeta(q_2, p_2)$.

To train the model with this objective, we use data from semantic textual similarity (STS) training datasets such as STS12~\cite{agirre2012semeval} and SICK~\cite{marelli-etal-2014-sick}. These datasets consist of triplets $(q_i, p_i, t_i) \in D$, where $(q_i, p_i)$ are text pairs and $t_i$ is the corresponding relevance score. A batch $B$ is constructed by selecting text values from a given number of triplets, with the ground truth similarity defined as $\zeta(q_i, p_i) = t_i$.

To enhance the model's performance across languages, we translate the STS12 and SICK datasets into multiple languages using machine translation models, i.e. WMT19~\cite{ng2019wmt19} and MADLAD-3B~\cite{kudugunta2023madlad}. While training on STS datasets is highly effective, obtaining large volumes of this data is challenging due to the human annotation required. As a result, we have incorporated various natural language inference (NLI) datasets into the training process.

During each training step, either an STS or NLI dataset is selected, and the batch is constructed solely from the chosen dataset, using the appropriate loss function. In other words, each batch contains text values from only one specific dataset.

For the relevant hyperparameters, see Appendix~\ref{table:hyperparameters}.

\subsubsection{Asymmetric Retrieval Adapters}
\label{sec:retrieval-adapter}

As discussed in Section~\ref{sec:rel:task-specific}, asymmetric retrieval tasks, such as question answering and traditional information retrieval, perform better with separate encoding mechanisms for queries and documents. In this work, we follow the method proposed by \citet{wang2022text}, using two distinct prefixes, but further separate the encoding processes by employing two specialized adapters, which are trained jointly. A detailed ablation study demonstrating the effectiveness of this approach is presented in Section~\ref{sec:eval:asymmetric-retrieval}.

Similar to previous works~\cite{wang2022text, li2023general, gunther2023jina2}, we use datasets containing hard negatives, such as MS-MARCO~\cite{bajaj2016ms} and Natural Questions (NQ)~\cite{47761}, to train the model to focus on subtle distinctions and to differentiate between relevant and similar but irrelevant documents. For retrieval training datasets without annotated negatives, we apply hard negative mining as outlined in~\citep{ren2021rocketqav2, wang2022text}, leveraging embedding models like BGE-large~\cite{xiao2023c} and BM25~\cite{robertson2009probabilistic}.

To incorporate the mined negatives into the training process, we apply the $\mathcal{L}_{\text{triplet}}$ loss function as seen in Equation~\eqref{eq:loss-multi-negatives}.


\subsubsection{Failure Analysis for Asymmetric Retrieval}

Since our \JEmbeddingVTwo{} models were trained on similar data to \JEmbeddingVThree{}, we conducted a failure analysis to identify issues common to models trained on these datasets. From this analysis, we identified the following points affecting retrieval tasks:

\begin{itemize}
    \item[F1.] \textbf{Misleading Syntactic Similarities:} Documents with high syntactic similarity to the query are often favored over gold/relevant documents with lower syntactic overlap.
    \item[F2.] \textbf{Misinterpretation of Named Entities:} Named entities are frequently not recognized as such, leading to documents being marked as relevant based on partial matches (e.g., "Sofia Albert" vs. "Albert Stone"). This occurs especially with proper nouns that have alternative, more common meanings (e.g., the novel title "The Company" vs. "the company").
    \item[F3.] \textbf{No Understanding of Polar Questions:} Complex yes-no (polar) questions are not handled effectively. As a result, the model retrieves documents with related content that do not necessarily answer the query.
    \item[F4.] \textbf{Preference for Low-Quality Documents:} \JEmbeddingVTwo{} and many other embedding models do not account for document quality, focusing solely on similarity and relevance. Consequently, low-quality documents (short, repetitive, or uninformative) that mention query terms are often retrieved but do not provide satisfactory answers.
\end{itemize}

To mitigate F1–F3, we crafted prompts to generate synthetic text examples targeting these specific failure cases. Each example consists of a query text, a preferred answer, and seven negative examples modeling the failure case.

For F4, we leveraged two preference learning datasets: \texttt{oasst1}\footnote{\url{https://huggingface.co/datasets/OpenAssistant/oasst1} (Accessed: 30-08-24)} and \texttt{oasst2}\footnote{\url{https://huggingface.co/datasets/OpenAssistant/oasst2} (Accessed: 30-08-24)} from the Open Assistant project~\cite{kopf2024openassistant}. These datasets contain questions and answers generated by LLMs with quality scores (0-1) based on human judgments.

We converted these datasets into hard negative training data by selecting queries with at least two answers. The highest-quality answer was treated as a positive match, while answers with at least a 0.3-point lower quality score were treated as negatives. If fewer than seven negatives were identified, additional negatives were randomly selected from other queries.

The effectiveness of training the retrieval adapter on this data is evaluated in Section~\ref{sec:eval:failure-cases}.

\subsubsection{Separation Adapter}
The separation adapter is designed to perform well on clustering and reranking tasks. It is trained to distinguish between text values belonging to the same group and those from different groups. For reranking tasks, the adapter separates relevant from irrelevant documents based on a query's information need. In clustering tasks, groups of texts are provided, and after calculating the embeddings and applying a clustering algorithm (e.g., k-means), the resulting clusters should reflect the correct groupings.

To train the adapter for this objective, we employ a variant of the CoSent loss $\mathcal{L}_{\mathrm{co}}$, introduced in Equation~\eqref{eq:cosentloss}. The training data consists of batches $B'$ made up of tuples $(x, l) \in B'$, where $x$ is a text value and $l$ is its label. To form a batch of text pairs compatible with $\mathcal{L}_{\mathrm{co}}$, we generate all pairwise combinations of text values that share the same label $l_i$ in the batch. The separation loss is then defined as follows:
\begin{flalign}
    &\mathcal{L}_{\mathrm{sep}}(B') := \mathcal{L}_{\mathrm{co}}(B) \nonumber\\
    &B = \{(x_i, x_j) | \exists l: (x_i, l), (x_j, l) \in B'\}
\end{flalign}
Since we have a limited amount of training data in this format, we observed that incorporating additional data from the pair training stage (Section \ref{sec:embedding-training}) into the training mix improves performance. We follow the same schema as used for the text matching adapter (Section~\ref{sec:text-matching}), where a specific dataset is sampled at each training step, and the corresponding loss function is applied.

Further details on the training data and hyperparameters are provided in Appendix~\ref{table:hyperparameters}.

\section{Evaluation}
\label{sec:evaluation}

In this section, we evaluate the performance of our model at various stages and conduct ablation studies on key architectural modifications. We begin by assessing the multilingual backbone model on a small subset of MTEB tasks in Section~\ref{sec:eval:backbone_performance}.

Next, we present a comprehensive evaluation of embedding tasks in Section~\ref{sec:mteb}, where our model is tested on a variety of MTEB tasks, both monolingual (English) and multilingual. Section~\ref{ssec:longembed} reports the model's performance on the LongEmbed MTEB evaluation, followed by an analysis of previously identified retrieval failure cases in Section~\ref{sec:eval:failure-cases}. Lastly, Section~\ref{sec:ablation-results} presents the ablation studies conducted on MRL and the retrieval adapter.

\begin{table}[t]
    \small
    \centering
    \setlength\tabcolsep{1.5pt}
    \begin{tabular}{lccc}
\toprule
Tasks & Jina-XLM-R & XLM-R & mBERT \\
\midrule
\multicolumn{4}{l}{\bfseries Multi-/Cross-lingual Tasks*} \\
STS22\textsuperscript{STS} & 64.52 & 64.80 & 60.59 \\
STS17\textsuperscript{STS} & 77.09 & 75.13 & 69.90 \\
PawsX\textsuperscript{PC} & 59.75 & 59.93 & 56.65 \\
\midrule
Average & \textbf{67.12} & 66.62 & 62.38 \\
\midrule
\midrule
\multicolumn{4}{l}{\bfseries EN Tasks} \\
QuoraRetrieval\textsuperscript{RT} & 84.88 & 83.49 & 77.81 \\
STS12\textsuperscript{STS} & 71.29 & 70.26 & 70.44 \\
STS13\textsuperscript{STS} & 81.55 & 80.89 & 75.69 \\
STS14\textsuperscript{STS} & 73.29 & 71.90 & 69.94 \\
STS15\textsuperscript{STS} & 83.17 & 81.03 & 77.99 \\
STS16\textsuperscript{STS} & 79.54 & 77.96 & 76.27 \\
STS17\textsuperscript{STS} & 84.48 & 84.00 & 80.35 \\
STS22\textsuperscript{STS} & 65.98 & 66.93 & 64.83 \\
TRECCOVID\textsuperscript{RT} & 60.28 & 42.96 & 35.31 \\
\midrule
Average & \textbf{76.05} & 73.38 & 69.18 \\
\bottomrule
\end{tabular}\\
    \small{*Results per task are averages across all languages\\ for that task.}
    \label{table:backbone_eval}
    \caption{Evaluation of multilingual pre-trained models after short embedding training on pair-wise data.}
    \label{tab:backbone_eval}
\end{table}
\begin{table*}[tbh]
\
    \centering
    \setlength{\tabcolsep}{4.5pt}
    \small{
        \begin{tabular}{lccccccccc}
 \toprule
  Model & Average & CF & CL & PC & RR & RT & STS & SM & \\
 \midrule %
     jina-embeddings-v2-base-en & 60.38 & 73.45 & 41.73 & 85.38 & 56.98 & 47.87 & 80.70 & \textbf{31.60} & \parbox[t]{2mm}{\multirow{5}{*}{\rotatebox[origin=c]{270}{\textbf{English}}}} \\
    jina-embeddings-v3 & \textbf{65.52} & \textbf{82.58} & 45.27 & 84.01 & 58.13 & 53.87 & \textbf{85.80} & 29.71 &  \\
    text-embedding-3-large & 64.60\textsuperscript{\#} & 75.45 & \textbf{49.01} & - & \textbf{59.16} & \textbf{55.44} & - & - & \\
    multilingual-e5-large-instruct & 64.41 & 77.56 & 47.10 & \textbf{86.19} & 58.58 & 52.47 & 84.78 & 30.39 & \\
    Cohere-embed-multilingual-v3.0 & 64.01 & 76.01 & 46.60 & 86.15 & 57.86 & 53.84 & 83.15 & 30.99 & \\
    \midrule
    jina-embeddings-v2-base-(zh/es/de)* & 60.54 & 65.69 & 39.36 & \textbf{82.95} & 66.57 & 58.24 & 66.60 & - & \parbox[t]{2mm}{\multirow{3}{*}{\rotatebox[origin=c]{270}{\textbf{Multi-L}}}} \\
    jina-embeddings-v3 & \textbf{64.44} & \textbf{71.46} & 46.71 & 76.91 & 63.98 & 57.98 & \textbf{69.83} & - &  \\
    multilingual-e5-large & 59.58 & 65.22 & 42.12 & 76.95 & 63.40 & 52.37 & 64.65 & - & \\
    multilingual-e5-large-instruct & 64.25 & 67.45 & \textbf{52.12} & 77.79 & \textbf{69.02} & \textbf{58.38}  & 68.77 & - & \\
 \bottomrule
\end{tabular}

        \hspace{100em}
        CF: Classification Accuracy \quad{}
        CL: Clustering $\mathcal{V}$ measure\quad{}
        PC: Pair Classification Average Precision\quad{} \\
        RR: Reranking MAP\quad{}
        RT: Retrieval nDCG@10\quad{}
        STS: Sentence Similarity Spearman Correlation\quad{} \\
        SM: Summarization Spearman Correlation\quad{} (Scores in \%)\\
        *: \JEmbeddingVTwo{} bilingual model suite, only evaluated on Chinese, Spanish, and German tasks.\\
        \#: The average MTEB score for \texttt{text-embedding-3-large} model as reported in the \href{https://openai.com/index/new-embedding-models-and-api-updates/}{announcement blog post}
    }
    \caption{Performance of multilingual text embedding models on MTEB tasks as averages.}
    \label{tab:mteb_results}
\end{table*}

\subsection{Performance of Jina-XLM-RoBERTa}
\label{sec:eval:backbone_performance}

We evaluate the Jina-XLM-RoBERTa model on a subset of English and multi-/cross-lingual MTEB tasks, conducting a comparative analysis against established multilingual models, specifically mBERT~\cite{devlin2019bert} and XLM-RoBERTa~\cite{conneau2020xlmr}, which are widely used as backbones for multilingual embedding models.

For this experiment, we follow the same training setup described in Section~\ref{sec:embedding-training}, but limit training to 1000 steps on a single GPU node, processing approximately 2 million pairs. Adapter tuning is excluded from this phase. As shown in Table~\ref{tab:backbone_eval}, our model outperforms both XLM-R and mBERT across all tasks, achieving an average of 76.05\% on monolingual English tasks and 67.12\% on multi-/cross-lingual tasks. These results reinforce our decision to continue training XLM-R for multilingual applications.
\subsection{Performance on MTEB}
\label{sec:mteb}

Table \ref{tab:mteb_results} summarizes the performance of various multilingual text embedding models across different MTEB tasks, divided into English and multilingual sections. \JEmbeddingVThree{} performs competitively, particularly in monolingual English tasks, where it achieves the highest Classification Accuracy (CF) score of \textbf{82.58} and the top Sentence Similarity (STS) score of \textbf{85.80}, demonstrating its robustness across both languages and tasks. Full evaluation results per task can be found in Appendix~\ref{table:english_mteb}. When averaging across all tasks, \JEmbeddingVThree{} scores 65.52, outperforming models such as \texttt{text-embedding-3-large}, \texttt{multilingual-e5-large-instruct}, and \texttt{Cohere-embed-multilingual-v3.0}. This indicates that \JEmbeddingVThree{} maintains strong monolingual English performance while being trained on a wide variety of languages.

Consulting the English MTEB leaderboard, we plotted the performance of the top 100 embedding models against their parameter sizes in Figure~\ref{img:scaling-law}. Notably, embedding models built on large language models (LLMs) perform only marginally better than \JEmbeddingVThree{} but entail significantly higher complexity and computational costs in real-world applications. For example, \texttt{e5-mistral-7b-instruct} achieves an average score of 66.63 across all 56 English MTEB tasks (approximately 1.03\% higher than \JEmbeddingVThree{}), but produces embeddings with a dimension of 4096 and has a parameter size of 7.1 billion. In contrast, \JEmbeddingVThree{} offers a more practical solution, with an embedding dimension of 1024 (or lower, using MRL with a manageable performance trade-off, as discussed in Section~\ref{sec:mlr-results}) and only 570 million parameters.

For multilingual performance, Table~\ref{tab:mteb_results} presents weighted averages across a wide selection of multilingual and cross-lingual MTEB tasks. The specific tasks, along with an explanation of which adapter was used for each task, are detailed in Appendix~\ref{table:multilingual_classification}, \ref{table:multilingual_PC}, \ref{table:multilingual_clustering}, \ref{table:multilingual_sts}, \ref{table:multilingual_reranking}, and \ref{table:multilingual_retrieval}. Note that \JEmbeddingVTwo{} refers to our bilingual model suite (\texttt{jina-embeddings-v2-base-zh}, \texttt{jina-embeddings-v2-base-es}, \texttt{jina-embeddings-v2-base-de}), which are evaluated only on Chinese, Spanish, and German monolingual and cross-lingual tasks, excluding all other tasks. Thus, the average score for \texttt{jina-embeddings-v2*} on the pair-classification tasks reflects performance on Chinese tasks alone. A complete report on pair-classification can be found in Appendix~\ref{table:multilingual_PC}. We do not report scores for \texttt{text-embedding-3-large} and \texttt{Cohere-embed-multilingual-v3.0} as these models were not evaluated on the full range of multilingual and cross-lingual MTEB tasks. However, our model outperforms \texttt{multilingual-e5-large} in all multilingual tasks except reranking and approaches the performance of \texttt{multilingual-e5-large-instruct}.

For \JEmbeddingVThree{}, all classification and pair-classification tasks are evaluated using the classification adapter, all STS tasks and three retrieval tasks (ArguAna, CQADupstackRetrieval, and QuoraRetrieval) are evaluated using the text matching adapter, all other retrieval tasks are evaluated using the retrieval adapter, and all clustering and reranking tasks are evaluated using the separation adapter.

\subsection{Performance on LongEmbed MTEB}
\label{ssec:longembed}

We have evaluated our model against \texttt{text-embedding-3-large}, \texttt{bge-m3}, and our previously released model suite \JEmbeddingVTwo{} on six long document retrieval tasks from the MTEB leaderboard. The results, presented in Table \ref{tab:mteb_long_embed}, demonstrate that \JEmbeddingVThree{} with the text-matching adapter achieves the highest average performance. These findings underscore the effectiveness of the RoPE-based positional embeddings, outperforming both the fixed positional embeddings used by \texttt{bge-m3} and the ALiBi-based approach employed in \JEmbeddingVTwo{}.

\begin{table*}[htb]
\
    \centering
    \setlength{\tabcolsep}{4.5pt}
    \small{
        \begin{tabular}{lcccccccc}
 \toprule
 Model & Average & NarrativeQA & Needle & Passkey & QMSum & SummScreen & WikiQA \\
 \midrule  
    jina-embeddings-v3* & \textbf{70.39} & 33.32 & \textbf{84.00} & \textbf{100.00} & \textbf{39.75} & 92.78 & 72.46 \\
    jina-embeddings-v2-base-en & 58.12 & 37.89 & 54.25 & 50.25 & 38.87 & 93.48 & 73.99 \\
    text-embedding-3-large & 51.30 & 44.09 & 29.25 & 63.00 & 32.49 & 84.80 & 54.16 \\
    baai-bge-m3 & 56.56 & \textbf{45.76} & 40.25 & 46.00 & 35.54 & \textbf{94.09} & \textbf{77.73} \\
 \bottomrule
 *: text-matching adapter
\end{tabular}

    }
    \caption{Evaluation of nDCG@10 [\%] on MTEB LongEmbed Tasks. }
    \label{tab:mteb_long_embed}
\end{table*}
\subsection{Retrieval Failures}
\label{sec:eval:failure-cases}
\begin{table}[htb]
\
    \centering
    \setlength{\tabcolsep}{4.5pt}
    \small{
\begin{tabular}{lcccc}
\toprule
\multicolumn{5}{l}{\textbf{Hand-Selected Failure Examples [mAP in \%]}} \\
\midrule
Model & F1 & F2 & F3 & F4 \\
\midrule
jina-embeddings-v2-base-en & 16.67 & 9.09 & 9.09 & 45.45\\
multilingual-e5-large & 46.97 & \textbf{45.45} & 27.27 & \textbf{81.82} \\
jina-embeddings-v3* & 46.97 & \textbf{45.45} & 27.27 & 9.09 \\
jina-embeddings-v3** & \textbf{62.12} & \textbf{45.45} & \textbf{45.45} & \textbf{81.82} \\
\midrule
\midrule
\multicolumn{5}{l}{\textbf{Synthetic Failure Examples [nDCG@10 in \%]}} \\
\midrule
Model & F1 & F2 & F3 & F4 \\
\midrule
jina-embeddings-v2-base-en & 37.83 & 93.84 & 46.14 & -- \\
multilingual-e5-large & 38.47 & 95.17 & 46.57 & -- \\
jina-embeddings-v3* & 37.63 & 93.04 & 45.91 & --\\
jina-embeddings-v3** & \textbf{45.78} & \textbf{98.78} & \textbf{47.62} & -- \\
 \bottomrule
\end{tabular}
\\\vspace{1mm}
F1: Misleading Syntactic Similarities, F2: Misinterpretation of Named Entities, F3: No Understanding of Polar Questions, F4 Preference for Low Quality Documents \\
*: pair training, **: retrieval adapter
}
\caption{Evaluation of Failure Cases }
\label{tab:failure-cases}
\end{table}
We conducted an analysis of retrieval failures typically observed when applying embedding models to retrieval tasks. This led to the identification of the four failure cases described in Section~\ref{sec:retrieval-adapter}. To assess whether training our retrieval adapter using synthetic and preference learning datasets mitigates these failures, we performed two quantitative evaluations.

The first experiment evaluated whether failure cases in existing retrieval benchmarks, such as HotpotQA~\cite{yang2018hotpotqa} and NaturalQuestions~\cite{47761}, were resolved. These examples\footnote{\url{https://huggingface.co/datasets/jinaai/retrieval-failure-examples}} consist of a query, a relevant document, and a less relevant document that is often assigned a higher retrieval score. Table~\ref{tab:failure-cases} presents the mean average precision (mAP) scores, showing that our model, after training with the retrieval adapters, handles these failure cases better or at least as effectively as our previously published \JEmbeddingVTwo{} models and the \texttt{multilingual-e5} model.

However, training on synthetic data does not seem to improve the model’s handling of failure case F2 (named entities), and failure cases F2 and F4 (low-quality documents) are handled equally well by the \texttt{multilingual-e5} model. Given the small size of the evaluation sets (fewer than 10 examples for most failure cases), we conducted a second analysis using a larger, synthetically generated set of challenging examples representing typical failures. In this case, failure case F4 was excluded, as LLMs are not suited to generating low-quality content due to their training on high-quality data.

In this second experiment, our model outperforms the other three models across all tasks, as shown in the lower part of Table~\ref{tab:failure-cases}. One limitation of this evaluation approach is that the synthetic examples may be too closely aligned with the training data, potentially making these failure cases easier for the model to resolve.

\subsection{Ablation Studies}
\label{sec:ablation-results}

\subsubsection{Matryoshka Representation Learning}
\label{sec:mlr-results}

Table \ref{tab:MRL_averages} presents the impact of MRL on the performance of our model. MRL is designed to improve the scalability and efficiency of text embeddings by enabling strong performance across a range of embedding dimensions (from 32 to 1024 in this case). The table is divided into two task categories: Retrieval and Semantic Textual Similarity (STS). A full evaluation can be found in Appendix~\ref{tab:MRL}.

In the Retrieval tasks reported in Appendix~\ref{tab:MRL}, our model consistently demonstrates strong performance across languages and datasets while using MRL, achieving competitive results even at lower-dimensional embeddings. For instance, in the XPQA Retrieval (French) task, the model reaches its highest score of 77.75 with a 1024-dimensional embedding, but also performs well with lower-dimensional embeddings, scoring 74.29 at 64 dimensions, representing only a 3.46\% decrease. This highlights the robustness of MRL in maintaining high performance without requiring the largest embedding size.

\begin{table}[]
    \centering
    \small{
    \begin{tabular}{lcc}
    \toprule
        MRL Dimension  & Retrieval & STS \\
    \midrule
        32      &   52.54 & 76.35 \\
        64      &   58.54 & 77.03 \\
        128     &   61.64 & 77.43 \\
        256     &   62.72 & 77.56 \\
        512     &   63.16 & 77.59 \\
        768     &   63.30 & 77.59 \\
        1024    &   63.35 & 77.58 \\
    \bottomrule
    \end{tabular}
    \caption{MRL ablation study on varying embedding sizes. Retrieval scores are nDCG@10 [\%], STS scores are spearman correlation coefficients [\%]. Full task list in Appendix~\ref{tab:MRL}.}
    \label{tab:MRL_averages}
    }
\end{table}

\subsubsection{Encoding Asymmetry for Retrieval}
\label{sec:eval:asymmetric-retrieval}
\begin{table}[h!]
\centering
    \setlength{\tabcolsep}{4.5pt}
    \small{
        \begin{tabular}{lcccc}
\toprule
\multirow{2}{*}{Task}& \multicolumn{2}{c}{w/ Instr.} & \multicolumn{2}{c}{w/o Instr.} \\ 
 & 1 Ad. & 2 Ad. & 1 Ad. & 2 Ad. \\ 
\midrule
SciFact          & 69.66 & 70.18  & \textbf{71.23}  & 70.58  \\ 
FiQA2018         & 46.70 & 47.21  & 47.41  & \textbf{47.49}  \\ 
TRECCOVID        & 73.70 & \textbf{75.05}  & 60.51  & 70.52  \\ 
NFCorpus         & 34.95 & \textbf{35.34}  & 34.57  & 34.76  \\ 
SCIDOCS          & 19.27 & 19.15  & \textbf{20.23}  & 19.03  \\ 
Touche2020       & 25.33 & 27.58  & 24.24  & \textbf{29.00}  \\ 
NarrativeQA      & 37.07 & 35.53  & \textbf{37.43}  & 35.33  \\ 
NQ               & 59.99 & 62.84  & 62.02  & \textbf{63.12}  \\ 
DBPedia          & 40.12 & \textbf{40.92}  & 37.64  & 40.71  \\ 
\midrule
Average              & 45.20 & \textbf{45.98}  & 43.92  & 45.62  \\ 
\bottomrule
\end{tabular}
    }
    \caption{Evaluation of Input Type Encoding for Asymmetric Retrieval Tasks [nDCG@10 \%]}
    \label{tab:retr_instr}
\end{table}
Table \ref{tab:retr_instr} provides key insights into the impact of using one versus two adapters, as well as the influence of instructions. For most tasks, the combination of two adapters with instructions resulted in the highest performance, demonstrating the advantages of increased model capacity and explicit guidance. For instance, the highest scores for TRECCOVID and NFCorpus were achieved with two adapters and instructions, scoring 75.05 and 35.34, respectively. In contrast, the absence of instructions generally led to reduced performance, particularly when using a single adapter. This trend underscores the importance of instructions in enhancing retrieval effectiveness.

On average, the use of two adapters consistently outperformed single adapters across both instruction settings, with average scores of 45.98 and 45.62, respectively, compared to 45.20 and 43.92 for single adapters. While instructions positively contributed to performance, the increased model capacity from using two adapters had a more significant impact. These findings suggest that for optimal performance in retrieval tasks, utilizing more adapters in conjunction with instructions is generally beneficial, though task-specific factors may influence the effectiveness of these configurations.

\section{Conclusion}
\label{sec:conclusion}

This paper introduces \JEmbeddingVThree{}, our latest text embedding model. By leveraging task-specific adapter tuning and failure-aware synthetic data augmentation on top of a robust backbone, \JEmbeddingVThree{} demonstrates competitive performance across a wide range of tasks. Extensive evaluations on both English and multilingual datasets highlight the model's strong performance while maintaining a reasonable parameter size.

We are particularly interested in evaluating and improving the model's performance on low-resource language and further analyzing systematic failures caused by low data availability. We plan to focus on this area going forward, further strengthening its capabilities in multilingual tasks where data availability is limited.

\FloatBarrier

\balance

\bibliographystyle{unsrtnat}
\bibliography{references}  

\clearpage
\pagenumbering{gobble}
\onecolumn

\appendix
\section{Appendix}
\setcounter{table}{0}
\renewcommand{\thetable}{A\arabic{table}}

\begin{figure*}[htbp]
\centering
\includegraphics[width=\linewidth]{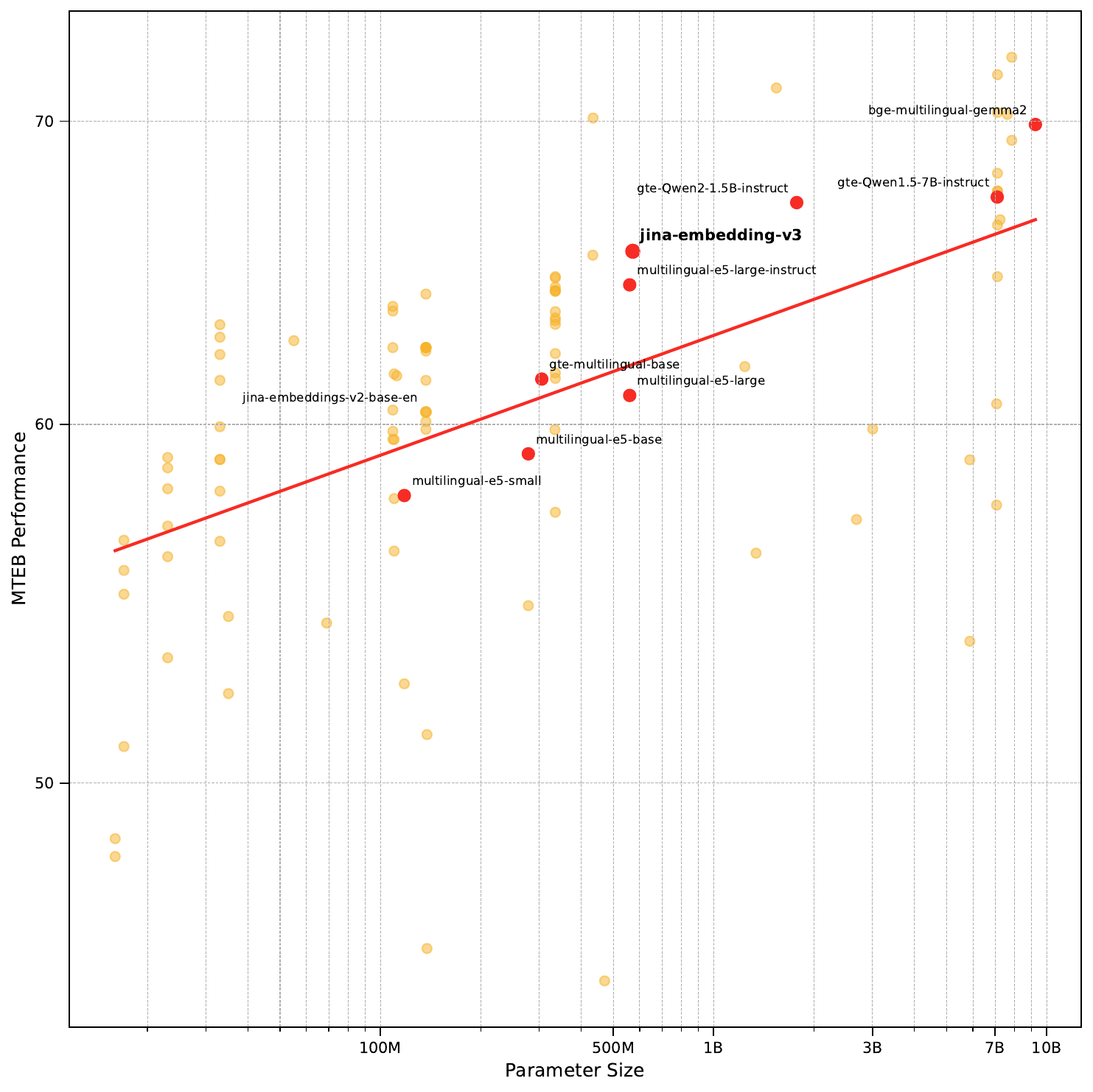}
\caption{Scaling law of embedding models. Each dot represents an embedding model. \JEmbeddingVThree{} demonstrates superior performance compared to models of similar size, showing a superlinear improvement over its predecessor, \JEmbeddingVTwo{}. This graph was created by selecting 100 embedding models from the MTEB leaderboard\footnote{\url{https://huggingface.co/spaces/mteb/leaderboard}}, excluding those without size information, typically closed-source or proprietary models. Submissions identified as outliers or trolling were also filtered out. The average MTEB performance on English tasks is plotted against the number of model parameters. The trendline, representing all models, is highlighted, with multilingual models emphasized in orange.}
\label{img:scaling-law}
\end{figure*}

\begin{table}[ht]
\centering
\small{
\begin{tabular}{l|cccccc}
\toprule
\bfseries Pre-Training & Number of Devices & Steps & Batch Size & Seq. Length & LR* & \\
\midrule
Pre-Training (Short Text) & $8$ & $100,000$ & $128 \times 8$ & 512 & $1\times{}10^{-4}$ & \\
Pre-Training (Long Text) & $8$ & $60,000$ & $8\times 8$ & $8192$ & $5\times{}10^{-5}$ & \\
\midrule
\bfseries Pair Training & Number of Devices & Steps & Batch Size & Seq. Length & LR* & $ \tau $\\
\midrule
Pair Training (Short Text) & 8 & 60,000 & $8 \times 2048$ & 192 & $3 \times 10^{-5} $ & $0.05$ \\
Pair Training (Long Text) & 2 & 50,000 & $2 \times 512$ & 1024 & $2 \times 10^{-5}$ & $0.02$ \\
\midrule
\bfseries Adapter Training & Number of Devices & Steps & Batch Size & Seq. Length & LR* & $ \tau $\\
\midrule
Retrieval Adapters & 1 & 20,000 & 128 & 512 & $5 \times 10^{-4}$ & 0.05 \\ 
Text Matching Adapter & 1 & 15,000 & 256 & 192 & $1 \times 10^{-4}$ & 0.05 \\ 
Classification Adapter & 1 & 11,500 & 256 & 192 & $5 \times 10^{-4}$ & 0.02 \\ 
Separation Adapter & 1 & 12,500 & 512 & 192 & $5 \times 10^{-5}$ & 0.02 \\ 
\bottomrule
\end{tabular}
}
\\ \vspace{1mm} * as we use a linear learning rate scheduler with warmup, this refers to the maximum learning rate
\caption{Hyperparameters}
\label{table:hyperparameters}
\end{table}

\begin{table*}[htb]
\
    \centering
    \setlength{\tabcolsep}{4.5pt}
    \small{
        \begin{tabular}{lccccccc}
\toprule
\textbf{Task} & \textbf{32} & \textbf{64} & \textbf{128} & \textbf{256} & \textbf{512} & \textbf{768} & \textbf{1024} \\
\midrule
\multicolumn{8}{c}{\textbf{Retrieval}} \\
\midrule
ArguAna (en) & 45.09 & 50.94 & 52.05 & 53.45 & 54.24 & 54.13 & 54.33 \\
QuoraRetrieval (en) & 86.21 & 87.95 & 88.63 & 88.92 & 88.99 & 89.03 & 89.09 \\
SciFact (en) & 55.07 & 64.94 & 69.95 & 71.52 & 72.47 & 72.05 & 72.31 \\
SCIDOCS (en) & 14.10 & 16.94 & 18.50 & 19.30 & 19.84 & 19.87 & 19.81 \\
TRECCOVID (en) & 65.74 & 73.02 & 76.92 & 77.07 & 76.04 & 77.36 & 77.72 \\
FiQA2018 (en) & 35.31 & 41.02 & 44.51 & 44.64 & 47.29 & 47.36 & 47.35 \\
NFCorpus (en) & 25.92 & 31.62 & 34.7 & 35.79 & 36.21 & 36.55 & 36.62 \\
AlloprofRetrieval (fr) & 40.19 & 46.91 & 51.22 & 53.35 & 53.98 & 54.39 & 54.28 \\
BSARDRetrieval (fr) & 15.79 & 19.85 & 23.55 & 24.48 & 24.92 & 25.15 & 24.76 \\
XPQARetrieval (fr) & 67.81 & 74.29 & 75.87 & 77.36 & 77.46 & 77.31 & 77.59 \\
CmedqaRetrieval (zh) & 28.67 & 32.20 & 34.16 & 35.32 & 35.85 & 35.82 & 35.93 \\
DuRetrieval (zh) & 73.65 & 79.36 & 81.66 & 82.83 & 83.11 & 83.15 & 83.13 \\
EcomRetrieval (zh) & 38.40 & 49.65 & 57.59 & 59.75 & 60.86 & 60.73 & 60.89 \\
MedicalRetrieval (zh) & 44.24 & 51.76 & 54.74 & 56.27 & 56.36 & 56.57 & 56.64 \\
GermanDPR (de) & 75.61 & 79.59 & 81.82 & 82.23 & 82.45 & 82.34 & 82.42 \\
GermanQuAD-Retrieval (de) & 90.00 & 93.13 & 94.78 & 95.29 & 95.47 & 95.48 & 95.46 \\
MLQARetrieval (es) & 51.95 & 59.66 & 64.66 & 67.49 & 68.32 & 68.61 & 68.63 \\
SpanishPassageRetrievalS2S (es) & 62.18 & 66.71 & 70.57 & 70.14 & 69.40 & 70.05 & 69.83 \\
XPQARetrieval (es) & 61.47 & 68.25 & 70.26 & 71.33 & 71.71 & 71.91 & 72.01 \\
RiaNewsRetrieval (ru) & 62.84 & 72.26 & 76.66 & 78.58 & 78.99 & 79.11 & 79.21 \\
RuBQRetrieval (ru) & 63.05 & 69.28 & 71.55 & 72.11 & 72.49 & 72.39 & 72.31 \\
\midrule
\multicolumn{8}{c}{\textbf{STS}} \\
\midrule
BIOSSES (en) & 87.75 & 88.00 & 88.04 & 88.04 & 88.45 & 88.66 & 88.69 \\
SICK-R (en) & 88.77 & 89.47 & 89.65 & 89.70 & 89.69 & 89.66 & 89.62 \\
STS12 (en) & 79.41 & 81.32 & 82.23 & 82.41 & 82.40 & 82.43 & 82.44 \\
STS15 (en) & 85.84 & 87.94 & 89.12 & 89.24 & 89.22 & 89.28 & 89.31 \\
STS22 (en) & 67.52 & 67.29 & 67.51 & 67.37 & 67.44 & 67.51 & 67.28 \\
STSBenchmarkMultilingualSTS (de) & 85.37 & 86.76 & 87.62 & 87.83 & 87.83 & 87.82 & 87.80 \\
STS22 (de) & 61.43 & 61.01 & 61.45 & 61.54 & 61.66 & 61.74 & 61.61 \\
STSBenchmarkMultilingualSTS (es) & 85.71 & 87.09 & 87.78 & 87.92 & 88.02 & 88.00 & 87.96 \\
STS22 (es) & 69.59 & 71.22 & 72.00 & 72.04 & 72.00 & 72.17 & 72.21 \\
STSBenchmarkMultilingualSTS (fr) & 85.80 & 87.00 & 87.35 & 87.41 & 87.42 & 87.43 & 87.44 \\
STS22 (fr) & 83.84 & 83.29 & 82.78 & 83.40 & 83.50 & 83.16 & 83.29 \\
SICKFr (fr) & 83.19 & 83.83 & 83.83 & 83.94 & 83.95 & 83.92 & 83.95 \\
STSBenchmarkMultilingualSTS (zh) & 84.73 & 85.33 & 85.66 & 85.69 & 85.72 & 85.69 & 85.64 \\
STS22 (zh) & 66.18 & 66.15 & 66.26 & 66.22 & 66.15 & 66.19 & 66.23 \\
AFQMC (zh) & 42.09 & 42.93 & 43.77 & 43.76 & 43.54 & 43.48 & 43.47 \\
STSBenchmarkMultilingualSTS (ru) & 84.81 & 85.69 & 86.01 & 86.30 & 86.34 & 86.30 & 86.30 \\
STS22 (ru) & 66.36 & 65.20 & 65.11 & 65.22 & 65.08 & 65.12 & 65.09 \\
STSBenchmarkMultilingualSTS (pl) & 82.55 & 83.99 & 85.18 & 85.52 & 85.58 & 85.50 & 85.48 \\
STS22 (pl) & 45.04 & 45.30 & 45.01 & 45.14 & 45.17 & 45.10 & 45.14 \\
CDSC-R (pl) & 91.03 & 91.87 & 92.33 & 92.48 & 92.59 & 92.62 & 92.63 \\
\bottomrule
\end{tabular}
    }
    \caption{MLR ablation study on retrieval and STS tasks}
    \label{tab:MRL}
\end{table*}
\begin{table*}[ht!]
    \centering
    \vspace{-1.3cm}
    \small{
        \begin{tabular}{lcccc}
\toprule
 & jina-embeddings-v3 & text-embedding-3 & multilingual-e5 & cohere-embed \\
 &  & -large & -large-instruct & -multilingual-v3 \\

\midrule
\bfseries Retrieval Average [nDCG@10 \%] & 53.87 & \textbf{55.44} & 52.47 & 53.84 \\
\hspace{0.5em}ArguAna (en) & 54.33 & 58.05\textsuperscript{$\star$} & 58.38 & 55.11 \\
\hspace{0.5em}CQADupstackRetrieval (en) & 42.36 & 47.54\textsuperscript{$\star$} & 42.71 & 40.64 \\
\hspace{0.5em}ClimateFEVER (en) & 42.36 & 30.27 & 29.86 & 29.96 \\
\hspace{0.5em}DBPedia (en) & 41.00 & 44.76 & 38.36 & 41.00 \\
\hspace{0.5em}FEVER (en) & 89.06 & 87.94 & 77.99 & 88.53 \\
\hspace{0.5em}FiQA2018 (en) & 47.35 & 55.00 & 47.71 & 44.10 \\
\hspace{0.5em}HotpotQA (en) & 64.67 & 71.58 & 69.32 & 70.61 \\
\hspace{0.5em}MSMARCO (en) & 40.82 & 40.24 & 40.43 & 43.45 \\
\hspace{0.5em}NFCorpus (en) & 36.62 & 42.07 & 35.53 & 36.42 \\
\hspace{0.5em}NQ (en) & 64.23 & 61.27 & 57.75 & 63.41 \\
\hspace{0.5em}QuoraRetrieval (en) & 89.09 & 89.05\textsuperscript{$\star$} & 89.15 & 88.92 \\
\hspace{0.5em}SCIDOCS (en) & 19.81 & 23.11 & 18.72 & 19.34 \\
\hspace{0.5em}SciFact (en) & 72.31 & 77.77 & 71.85 & 70.05 \\
\hspace{0.5em}Touche2020 (en) & 26.30 & 23.35 & 27.25 & 32.70 \\
\hspace{0.5em}TRECCOVID (en) & 77.72 & 79.56 & 82.00 & 83.37 \\
\midrule
\bfseries Clustering Average [v--measure] & 45.27 & \textbf{49.00} & 47.10 & 46.60 \\
\hspace{0.5em}ArxivClusteringP2P (en) & 46.66 & 49.01 & 46.40 & 48.16 \\
\hspace{0.5em}ArxivClusteringS2S (en) & 39.27 & 44.45 & 40.49 & 40.79 \\
\hspace{0.5em}BiorxivClusteringP2P (en) & 38.91 & 38.03 & 40.94 & 40.50 \\
\hspace{0.5em}BiorxivClusteringS2S (en) & 34.42 & 36.53 & 36.28 & 36.91 \\
\hspace{0.5em}MedrxivClusteringP2P (en) & 34.80 & 32.70 & 36.93 & 36.18 \\
\hspace{0.5em}MedrxivClusteringS2S (en) & 32.42 & 31.27 & 35.54 & 33.44 \\
\hspace{0.5em}RedditClustering (en) & 55.40 & 67.84 & 56.60 & 58.11 \\
\hspace{0.5em}RedditClusteringP2P (en) & 62.87 & 67.96 & 64.27 & 65.02 \\
\hspace{0.5em}StackExchangeClustering (en) & 65.66 & 76.26 & 66.85 & 68.12 \\
\hspace{0.5em}StackExchangeClusteringP2P (en) & 35.08 & 36.88 & 42.46 & 35.22 \\
\hspace{0.5em}TwentyNewsgroupsClustering (en) & 52.49 & 58.14 & 51.33 & 50.14 \\
\midrule
\bfseries Reranking Average [map] & 58.13 & \textbf{59.16} & 58.58 & 57.86 \\
\hspace{0.5em}AskUbuntuDupQuestions & 65.04 & 65.03 & 63.89 & 62.13 \\
\hspace{0.5em}MindSmallReranking & 30.83 & 29.86 & 33.09 & 32.59 \\
\hspace{0.5em}SciDocsRR & 84.88 & 86.66 & 85.87 & 84.31 \\
\hspace{0.5em}StackOverflowDupQuestions & 51.77 & 55.08 & 51.45 & 52.40 \\
\midrule
\bfseries Classification Average [acc.]  & \textbf{82.58} & 75.45 & 77.56 & 76.01 \\
\hspace{0.5em}AmazonCounterfactualClassification & 89.49 & 78.83 & 76.24 & 77.85 \\
\hspace{0.5em}AmazonPolarityClassification & 95.38 & 92.85 & 96.29 & 95.60 \\
\hspace{0.5em}AmazonReviewsClassification & 49.77 & 48.70 & 56.72 & 49.79 \\
\hspace{0.5em}Banking77Classification (en) & 84.08 & 85.69 & 85.73 & 86.09 \\
\hspace{0.5em}EmotionClassification (en) & 73.30 & 51.58 & 51.51 &  48.15\\
\hspace{0.5em}ImdbClassification (en) & 91.90 & 87.67 & 94.60 & 93.97 \\
\hspace{0.5em}MTOPDomainClassification (en) & 97.49 & 95.36 & 93.93 & 94.92 \\
\hspace{0.5em}MTOPIntentClassification (en) & 84.53 & 75.05 & 82.46 & 78.89 \\
\hspace{0.5em}MassiveIntentClassification (en) & 77.60 & 74.64 & 77.06 & 74.51 \\
\hspace{0.5em}MassiveScenarioClassification (en) & 84.71 & 79.79 & 80.47 & 79.00 \\
\hspace{0.5em}ToxicConversationsClassification (en) & 91.28 & 72.92 & 71.06 & 71.20 \\
\hspace{0.5em}TweetSentimentExtractionClassification (en) & 71.39 & 62.22 & 64.62 & 62.18 \\
\midrule
\bfseries Pair--Classification Average [AveP]  & 84.01 & -- & \textbf{86.19} & 86.15 \\
\hspace{0.5em}SprintDuplicateQuestions (en) & 96.98 & -- & 91.18 & 96.79 \\
\hspace{0.5em}TwitterSemEval2015 (en) & 70.91 & -- & 80.27 & 75.16 \\
\hspace{0.5em}TwitterURLCorpus (en) & 84.13 & -- & 87.12 & 86.49 \\
\midrule
\bfseries STS Average [Spearman]  & \textbf{85.80} & -- & 84.78 & 83.15 \\
\hspace{0.5em}BIOSSES & 88.69 & -- & 86.96 & 85.01 \\
\hspace{0.5em}SICK--R & 89.62 & -- & 81.73 & 82.18 \\
\hspace{0.5em}STS12 & 82.44 & -- & 82.57 & 77.62 \\
\hspace{0.5em}STS13 & 89.49 & -- & 87.15 & 85.16 \\
\hspace{0.5em}STS14 & 84.94 & -- & 84.97 & 80.02 \\
\hspace{0.5em}STS15 & 89.31 & -- & 91.05 & 88.92 \\
\hspace{0.5em}STS16 & 86.84 & -- & 87.31 & 86.92 \\
\hspace{0.5em}STS17 (en--en) & 89.97 & -- & 90.03 & 90.09 \\
\hspace{0.5em}STS22 (en) & 67.27 & -- & 67.63 & 66.81 \\
\hspace{0.5em}STSBenchmark & 89.44 & -- & 88.38 & 88.79 \\
\midrule
\bfseries Summarization Average [Spearman]  & 29.7 & -- & 30.39 & \textbf{30.99}  \\
\hspace{0.5em}SummEval  & 29.70 & -- & 30.39 & 30.99 \\
\bottomrule
\end{tabular}

        \begin{minipage}{0.89\textwidth}\raggedright
        \vspace{2mm}
        \end{minipage}
        \textsuperscript{$\star$}: ArguAna, CQADupstackRetrieval and QuoraRetrieval are evaluated using the text matching adapter for \JEmbeddingVThree{}
    }
    \caption{Performance of embedding models on English MTEB tasks.}
    \label{table:english_mteb}
\end{table*}

\begin{table*}[ht!]
    \centering
    \vspace{-1.3cm}
    \small{
        \begin{tabular}{lcccccc}
\toprule
 & je-v2* & je-v3 & m-e5-large & te-3-large & Cohere-embed & m-e5-instr \\
 \midrule
\bfseries Retrieval Average [nDCG@10] & 58.24 & 57.98 & 52.37 &  &  & \textbf{58.38} \\
\midrule
\bfseries MIRACL (all) & -- & 61.9 & \textbf{66.5} & 54.9 & 52.8 & 65.7 \\
\midrule
\bfseries Chinese Average & \textbf{69.40} & 68.60 & 63.66 & -- & -- & 64.94 \\
\hspace{0.5em}CmedqaRetrieval & 39.15 & 35.93 & 28.67 & -- & -- & 34.11 \\
\hspace{0.5em}CovidRetrieval & 81.22 & 78.92 & 75.51 & -- & -- & 75.76 \\
\hspace{0.5em}DuRetrieval & 84.57 & 83.13 & 85.32 & -- & -- & 85.14 \\
\hspace{0.5em}EcomRetrieval & 63.95 & 60.89 & 54.75 & -- & -- & 53.94 \\
\hspace{0.5em}MedicalRetrieval & 57.12 & 56.64 & 51.44 & -- & -- & 56.56 \\
\hspace{0.5em}MMarcoRetrieval & 77.96 & 79.69 & 79.2 & -- & -- & 78.82 \\
\hspace{0.5em}T2Retrieval & 80.59 & 83.16 & 76.11 & -- &-- & 82.92 \\
\hspace{0.5em}VideoRetrieval & 70.62 & 70.43 & 58.25 & -- & -- & 52.28 \\
\midrule
\bfseries German Average & 38.73 & \textbf{42.32} & 35.63 & -- & -- & 38.19 \\
\hspace{0.5em}GerDaLIR & 17.20 & 16.23 & 6.53 & -- & -- & 9.33 \\
\hspace{0.5em}GermanDPR & 79.50 & 82.42 & 82.89 & -- & -- & 80.87 \\
\hspace{0.5em}XMarket & 19.50 & 28.32 & 17.46 & -- & -- & 24.37 \\
\midrule
\bfseries Spanish Average & \textbf{52.12} & 47.75 & 44.55  & -- & -- & 47.86 \\
\hspace{0.5em}MintakaRetrieval & 28.30 & 26.91 & 33.23  & -- & -- & 34.50 \\
\hspace{0.5em}XMarket & 19.70 & 26.59 & 13.48  & -- & -- & 25.00 \\
\hspace{0.5em}SpanishPassageRetrievalS2S & 81.07 & 69.83 & 72.32  & -- & -- & 71.49 \\
\hspace{0.5em}SpanishPassageRetrievalS2P & 66.15 & 43.4 & 41.96  & -- & -- & 43.17 \\
\hspace{0.5em}XPQARetrieval & 65.40 & 72.00 & 61.77  & -- & -- & 65.12 \\
\midrule
\bfseries French Average & -- & 53.48 & 42.17 & -- & 40.42 & \textbf{62.46} \\
\hspace{0.5em}AlloprofRetrieval & -- & 54.28 & 38.15  & -- & 38.36 & 52.07 \\
\hspace{0.5em}BSARDRetrieval & -- & 24.76 & 0.27  & -- & 0.14 & 66.21 \\
\hspace{0.5em}MintakaRetrieval (fr) & -- & 26.91 & 25.20  & -- & 25.44 & 33.49 \\
\hspace{0.5em}SyntecRetrieval & -- & 83.85 & 81.07  & -- & 79.27 & 87.78 \\
\hspace{0.5em}XPQARetrieval (fr) & -- & 77.58 &   66.15 & -- & 58.87 & 72.73 \\
\midrule
\bfseries Russian Average & -- & 75.76 & 77.39  & -- & -- & \textbf{78.57} \\
\hspace{0.5em}RiaNewsRetrieval (rus--Cyrl) & -- & 79.21 & 80.67  & -- & -- & 83.25 \\
\hspace{0.5em}RuBQRetrieval (rus--Cyrl) & -- & 72.30 & 74.11  & -- & -- & 73.88 \\
\bottomrule
\end{tabular}
        \begin{minipage}{0.89\textwidth}\raggedright
        \vspace{2mm}
        \end{minipage}
        \hspace{100em}
        je-v2*: jina-embeddings-v2-base-(zh/de/es) bilingual model suite \quad{}
        je-v3: jina-embeddings-v3 \\
        m-e5-large: multilingual-e5-large \quad{}
        te-3-large: text-embedding-3-large \\
        Cohere-embed: Cohere-embed-multilingual-v3 \quad{}
        m-e5-instr: multilingual-e5-large-instruct \\
    }
    \caption{Performance of bi-/multilingual models on multilingual retrieval tasks.}
    \label{table:multilingual_retrieval}
\end{table*}
\begin{table*}[ht]
    \centering
    \vspace{-1.3cm}
    \small{
        \begin{tabular}{lcccc}
\toprule
 & je-v2* & je-v3 & m-e5-large & m-e5-instr \\
 \midrule
\bfseries STS Average [Spearman \%] & 66.60 & \textbf{69.83} & 64.65 & 68.77 \\
\midrule
\bfseries Chinese Average & \textbf{59.38} & 54.18 & 48.29 & 53.63 \\
\hspace{0.5em}AFQMC & 50.59 & 43.47 & 33.02 &  37.54 \\
\hspace{0.5em}ATEC & 51.28 & 49.06 & 39.81 & 43.26 \\
\hspace{0.5em}BQ & 66.07 & 63.2 & 46.44 & 48.80 \\
\hspace{0.5em}LCQMC & 75.74 & 75.95 & 75.95 & 76.05 \\
\hspace{0.5em}PAWSX & 41.48 & 15.56 & 14.63 & 15.02 \\
\hspace{0.5em}QBQTC & 38.11 & 36.43 & 29.77 & -- \\
\hspace{0.5em}STS22 & 69.25 & 66.23 & 65.64 & 73.10 \\
\hspace{0.5em}STSB & 82.55 & 83.54 & 81.08 & 81.67 \\
\midrule
\bfseries German Average & 78.47 & \textbf{78.97} & 74.83 &  77.22 \\
\hspace{0.5em}STSBenchmarkMultilingualSTS & 88.45 & 87.8 & 84.27 & 85.37 \\
\hspace{0.5em}GermanSTSBenchmark & 88.32 & 87.5 & 83.64 & 84.84 \\
\hspace{0.5em}STS22  & 58.63 & 61.61 & 56.59 & 61.45 \\
\midrule
\bfseries Spanish Average & 77.72 & \textbf{80.09} & 74.21 & 79.93 \\
\hspace{0.5em}STSBenchmarkMultilingualSTS & 86.38 & 87.96 & 83.81 & 86.15 \\
\hspace{0.5em}STS22 & 69.06 & 72.21 & 64.60 & 73.70 \\
\midrule
\bfseries French Average & -- & \textbf{84.89} & 79.37 & 82.62 \\
\hspace{0.5em}STSBenchmarkMultilingualSTS & -- & 87.44 & 76.79 & 84.93 \\
\hspace{0.5em}STS22 & -- & 83.29 & 82.53 & 82.73 \\	
\hspace{0.5em}SICKFr & -- & 83.95 & 78.78 & 80.20 \\
\midrule
\bfseries Polish Average & -- & \textbf{72.76} & 70.45 & 72.15 \\
\hspace{0.5em}CDSC--R & -- & 92.63 & 91.00 & 92.35 \\
\hspace{0.5em}SICK--R--PL & -- & 80.53 & 75.08 & 77.61 \\
\hspace{0.5em}STS22 & -- & 45.14 & 34.66 & 46.49 \\
\midrule
\bfseries Russian Average & -- & \textbf{81.5} & 74.48 & 79.67 \\
\hspace{0.5em}RUParaPhraserSTS & -- & 76.77 & 71.82 & 75.37 \\
\hspace{0.5em}RuSTSBenchmarkSTS & -- & 86.22 & 83.15 & 83.97 \\
\bottomrule
\end{tabular}

        \begin{minipage}{0.89\textwidth}\raggedright
        \vspace{2mm}
        \end{minipage}
        \hspace{100em}
        je-v2*: jina-embeddings-v2-base-(zh/de/es) bilingual model suite \quad{}
        je-v3: jina-embeddings-v3 \\
        m-e5-large: multilingual-e5-large \quad{}
        m-e5-instr: multilingual-e5-large-instruct \\
    }
    \caption{Performance of bi-/multilingual models on STS tasks.}
    \label{table:multilingual_sts}
\end{table*}
\begin{table*}[ht]
    \centering
    \vspace{-1.3cm}
    \small{
        \begin{tabular}{lcccc}
\toprule
 & je-v2* & je-v3 & m-e5-large & m-e5-instr \\
 \midrule
\bfseries Pair Classification Average [AveP]  & \textbf{82.95} & 76.91 & 76.95 & 77.79 \\
\midrule
\bfseries Chinese Average & \textbf{82.95} & 72.76 & 69.89 & 66.54 \\
\hspace{0.5em}Cmnli & 85.27 & 79.10 & 78.18 & 71.41 \\
\hspace{0.5em}Ocnli & 80.62 & 66.42 & 61.6 & 61.67 \\
\midrule
\bfseries French Average & -- & 76.57 & 76.19 & \textbf{77.23} \\
\hspace{0.5em}OpusparcusPC & -- & 93.74 & 93.89 & 94.72 \\
\hspace{0.5em}PawsXPairClassification & -- & 59.4 & 58.5 & 59.73 \\
\midrule
\bfseries Polish Average & -- & 83.70 & 85.5 & \textbf{87.16} \\
\hspace{0.5em}CDSC--E &--  & 73.02 & 74.47 & 76.18 \\
\hspace{0.5em}PpcPC & -- & 91.43 & 92.18 & 93.45 \\
\hspace{0.5em}PSC & -- & 99.19 & 99.39 & 99.31 \\
\hspace{0.5em}SICK--E--PL & -- & 71.15 & 75.96 & 79.68 \\
\midrule
\bfseries Russian Average & -- & 58.77 & 58.4 & \textbf{63.92} \\
\hspace{0.5em}TERRa & -- & 58.77 & 58.4 & 63.92 \\
\bottomrule
\end{tabular}

        \begin{minipage}{0.89\textwidth}\raggedright
        \vspace{2mm}
        \end{minipage}
        \hspace{100em}
        je-v2*: jina-embeddings-v2-base-(zh/de/es) bilingual model suite \quad{}
        je-v3: jina-embeddings-v3 \\
        m-e5-large: multilingual-e5-large \quad{}
        m-e5-instr: multilingual-e5-large-instruct \\
    }
    \caption{Performance of bi-/multilingual models on pair-classification tasks.}
    \label{table:multilingual_PC}
\end{table*}
\begin{table*}[ht]
    \centering
    \vspace{-1.3cm}
    \small{
        \begin{tabular}{lcccc}
\toprule
 & je-v2* & je-v3 & m-e5-large & m-e5-instr \\
 \midrule
\bfseries Classification Average [acc.]  & 65.69 & \textbf{71.46}  & 65.22 & 67.45 \\
\midrule
\bfseries Chinese Average & 64.94 & \textbf{69.07} & 67.34 & 67.85 \\
\hspace{0.5em}AmazonReviews  & 34.94 & 44.77 & 38.83 & 45.11 \\
\hspace{0.5em}IFlyTek & 47.36 & 41.68 & 45.47 & 44.06 \\
\hspace{0.5em}JDReview & 79.57 & 83.51 &  80.99 & 80.21 \\
\hspace{0.5em}MassiveIntent & 68.2 & 73.22 & 71.12 & 67.85 \\
\hspace{0.5em}MassiveScenario & 71.93 & 80.82 & 76.83 & 72.43 \\
\hspace{0.5em}MultilingualSentiment & 63.29 & 73.21 &  68.58 & 72.44 \\
\hspace{0.5em}OnlineShopping & 87.00 & 91.88 & 90.81 & 91.83 \\
\hspace{0.5em}TNews & 47.65 & 45.97 &  48.38 & 49.84 \\
\hspace{0.5em}Waimai & 84.54 & 86.59 &  85.02 & 86.85 \\
\midrule
\bfseries German Average & 	65.66 &  \textbf{79.03} & 67.77 & 69.55 \\
\hspace{0.5em}AmazonCounterfactual & 68.92 & 89.61 &  68.66 & 65.63 \\
\hspace{0.5em}AmazonReviews & 37.72 & 48.07 &  46.50 & 54.54 \\
\hspace{0.5em}MassiveIntent & 63.89 & 74.84 &  63.82 & 65.90 \\
\hspace{0.5em}MassiveScenario & 71.25 & 83.97 &  71.25 & 72.69 \\
\hspace{0.5em}MTOPDomain & 88.37 & 96.89 &  90.44 & 90.00 \\
\hspace{0.5em}MTOPIntent & 63.83 & 80.79 &  65.97 & 68.55 \\
\midrule
\bfseries Spanish Average & 67.10 &  \textbf{77.59} & 65.33 & 66.89 \\
\hspace{0.5em}AmazonReviews & 38.68 & 48.10 &  44.35 & 49.88 \\
\hspace{0.5em}MassiveIntent & 66.93 & 75.32 &   64.01 & 65.57 \\
\hspace{0.5em}MassiveScenario & 71.23 & 82.58 &  69.07 & 70.01 \\
\hspace{0.5em}MTOPDomain & 89.89 & 97.17 & 88.34 & 89.12 \\
\hspace{0.5em}MTOPIntent & 68.76 & 84.77 &  61.90 & 69.86 \\
\midrule
\bfseries French Average & -- & \textbf{76.70} & 68.39 & 69.32 \\
\hspace{0.5em}AmazonReviews & -- & 47.3  & 41.91 & 49.78\\
\hspace{0.5em}MasakhaNEWS & -- & 74.99 & 79.38 & 78.93 \\
\hspace{0.5em}MassiveIntent & -- & 76.33  & 69.34 &  66.88 \\
\hspace{0.5em}MassiveScenario & --  & 83.09 & 73.87 &  71.16 \\
\hspace{0.5em}MTOPDomain & -- & 96.30 &  86.41 & 85.89 \\
\hspace{0.5em}MTOPIntent & -- &  82.21 & 59.43 &  63.29 \\
\midrule
\bfseries Polish Average & -- & \textbf{70.81} & 63.82 & 65.99 \\
\hspace{0.5em}AllegroReviews & -- & 49.51  & 41.14 & 45.32 \\
\hspace{0.5em}CBD & -- & 69.99 & 69.90 & 74.25 \\
\hspace{0.5em}MassiveIntent & -- & 75.37 & 65.07 & 67.45 \\
\hspace{0.5em}MassiveScenario & --  & 82.1 &  69.82 &  71.44 \\
\hspace{0.5em}PAC & -- & 68.4 & 70.37 & 65.69 \\
\hspace{0.5em}PolEmo2.0--IN & -- &  83.75 & 77.06 & 80.99 \\
\hspace{0.5em}PolEmo2.0--OUT & -- & 66.53  & 53.38 & 56.84 \\
\midrule
\bfseries Russian Average & -- & 59.84 & 58.92 & \textbf{65.38} \\
\hspace{0.5em}eoreviewClassification & -- & 48.01 & 49.69 & 55.92 \\
\hspace{0.5em}HeadlineClassification & -- & 75.08 & 77.19 & 86.18 \\
\hspace{0.5em}Inappropriateness & -- & 61.05 & 61.60 & 65.61 \\
\hspace{0.5em}RuSciBenchGRNTI & -- & 59.19 & 58.20 & 65.08 \\
\hspace{0.5em}RuSciBenchOECD & -- & 45.56 & 43.91 & 50.18 \\
\hspace{0.5em}Kinopoisk & -- & 62.39 & 56.59 & 66.12 \\
\hspace{0.5em}RuReviews & -- & 67.58 & 65.28 & 68.54 \\
\bottomrule
\end{tabular}
        \begin{minipage}{0.89\textwidth}\raggedright
        \vspace{2mm}
        \end{minipage}
        \hspace{100em}
        je-v2*: jina-embeddings-v2-base-(zh/de/es) bilingual model suite \quad{}
        je-v3: jina-embeddings-v3 \\
        m-e5-large: multilingual-e5-large \quad{}
        te-3-large: text-embedding-3-large \\
        Cohere-embed: Cohere-embed-multilingual-v3 \quad{}
        m-e5-instr: multilingual-e5-large-instruct \\
    }
    \caption{Performance of bi-/multilingual models on classification tasks.}
    \label{table:multilingual_classification}
\end{table*}
\begin{table*}[ht]
    \centering
    \vspace{-1.3cm}
    \small{
        \begin{tabular}{lcccccc}
\toprule
 & je-v2* & je-v3 & m-e5-large & te-3-large & m-e5-instr \\
 \midrule
\bfseries Clustering Average [v--measure] & 39.36 & 46.71 & 42.12 & 46.65 & \textbf{52.12} \\
\midrule
\bfseries Chinese Average & 46.47 & 50.22 & 48.23 & 48.75 & \textbf{52.72} \\
\hspace{0.5em}CLSClusteringS2S & 38.4 & 38.79 & 38.59 & 38.82 & 40.65 \\
\hspace{0.5em}CLSClusteringP2P & 39.97 & 39.96 & 40.68 & 39.64 & 42.28 \\
\hspace{0.5em}ThuNewsClusteringS2S & 53.42 & 58.98 & 55.59 & 55.40 & 60.81 \\
\hspace{0.5em}ThuNewsClusteringP2P & 54.08 & 63.13 & 58.05 & 61.15 & 67.12 \\
\midrule
\bfseries German Average & 29.91 & 36.54 & 34.60 & 37.26 & \textbf{39.73} \\
\hspace{0.5em}TenKGnadClusteringS2S & 25.01 & 39.76 & 33.9 & 37.25 & 41.42 \\
\hspace{0.5em}TenKGnadClusteringP2P & 43.07 & 43.76 & 45.8 & 45.03 & 52.76 \\
\hspace{0.5em}BlurbsClusteringS2S & 16.67 & 21.00 & 16.51 & 21.98 & 20.99 \\
\hspace{0.5em}BlurbsClusteringP2P & 34.89 & 41.63 & 42.19 & 44.78 & 43.78 \\
\midrule
\bfseries Spanish Average & 44.05 & 45.09 & 41.52 & \textbf{48.59} & 46.50 \\
\hspace{0.5em}SpanishNewsClusteringP2P & 48.31 & 45.04 & 41.66 & 52.96 & 45.98 \\
\hspace{0.5em}FloresClusteringS2S  & 39.79 & 45.13 & 41.38 & 44.21 & 47.01 \\
\midrule
\bfseries French Average & -- & 44.95 & 38.63 & 45.79 & \textbf{55.80} \\
\hspace{0.5em}HALClusteringS2S &--  & 27.48 & 22.44 & 27.87 & 28.25 \\
\hspace{0.5em}AlloProfClusteringS2S & -- & 46.86 & 32.26 & 53.56 & 60.03 \\
\hspace{0.5em}AlloProfClusteringP2P & -- & 63.88 & 62.99 & 62.72 & 68.77 \\
\hspace{0.5em}MLSUMClusteringP2P & -- & 44.69 & 44.04 & 45.01 & 47.25 \\
\hspace{0.5em}MLSUMClusteringS2S & -- & 44.88 & 37.65 & 38.41 & 46.18 \\
\hspace{0.5em}MasakhaNEWSClusteringP2P & -- &  42.36& 40.49 & 53.23 & 70.48 \\
\hspace{0.5em}MasakhaNEWSClusteringS2S & -- & 44.48 & 30.56 & 39.71 & 69.65 \\
\midrule
\bfseries Russian Average & -- & 60.80 & 52.55 & 57.10 & \textbf{63.04} \\
\hspace{0.5em}GeoreviewClusteringP2P  & -- & 73.62 & 60.51 & 71.59 & 74.29 \\
\hspace{0.5em}RuSciBenchGRNTIClusteringP2P  & -- & 57.99 & 52.03 & 53.69 & 61.74 \\
\hspace{0.5em}RuSciBenchOECDClusteringP2P  & -- & 50.79 & 45.11 & 46.01 & 53.09 \\
\bottomrule
\end{tabular}

        \begin{minipage}{0.89\textwidth}\raggedright
        \vspace{2mm}
        \end{minipage}
        \hspace{100em}
        je-v2*: jina-embeddings-v2-base-(zh/de/es) bilingual model suite \quad{}
        je-v3: jina-embeddings-v3 \\
        m-e5-large: multilingual-e5-large \quad{}
        te-3-large: text-embedding-3-large \\
        m-e5-instr: multilingual-e5-large-instruct \\
    }
    \caption{Performance of bi-/multilingual models on clustering tasks.}
    \label{table:multilingual_clustering}
\end{table*}
\begin{table*}[ht]
    \centering
    \vspace{-1.3cm}
    \small{
        \begin{tabular}{lccccc}
\toprule
 & je-v2* & je-v3 & m-e5-large & m-e5-instr \\
 \midrule
\bfseries Reranking Average & 66.57 & 63.98 & 63.40 & \textbf{69.02} \\
\midrule
\bfseries Chinese Average & \textbf{66.57} & 60.64 & 56.00 & 60.68 \\
\hspace{0.5em}CMedQAv1 & 83.64 & 77.96 & 68.25 & 75.94 \\
\hspace{0.5em}CMedQAv2 & 83.74 & 78.22 & 68.56 & 76.10 \\
\hspace{0.5em}MMarcoReranking & 31.54 & 21.05 & 21.34 & 23.59 \\
\hspace{0.5em}T2Reranking & 67.37 & 65.31 & 65.83 & 67.10 \\
\midrule
\bfseries French Average & -- & 69.86 & 72.14 & \textbf{82.30} \\
\hspace{0.5em}AlloprofReranking & -- & 66.39 & 57.37 & 74.65 \\
\hspace{0.5em}SyntecReranking & -- & 73.32 & 86.9 & 89.95 \\
\midrule
\bfseries Russian Average & -- & 65.57 & 75.58 & \textbf{75.85} \\
\hspace{0.5em}RuBQReranking  & -- & 65.57 &  75.58 & 75.85 \\
\bottomrule
\end{tabular}

        \begin{minipage}{0.89\textwidth}\raggedright
        \vspace{2mm}
        \end{minipage}
        \hspace{100em}
        je-v2*: jina-embeddings-v2-base-(zh/de/es) bilingual model suite \quad{}
        je-v3: jina-embeddings-v3 \\
        m-e5-large: multilingual-e5-large \quad{}
        m-e5-instr: multilingual-e5-large-instruct \\
    }
    \caption{Performance of bi-/multilingual models on reranking tasks.}
    \label{table:multilingual_reranking}
\end{table*}
%





\end{document}